\documentclass{article}

\IfFileExists{arxiv.sty}{
  \usepackage{arxiv}
}{}
\usepackage{multirow}
\usepackage[utf8]{inputenc} 
\usepackage[T1]{fontenc}
\usepackage{textcomp}
\usepackage{hyperref}       
\usepackage{url}            
\usepackage{booktabs}       
\usepackage{amsfonts}       
\usepackage{nicefrac}       
\usepackage{microtype}      
\usepackage{lipsum}
\usepackage{graphicx}
\usepackage{xcolor,colortbl}
\usepackage{pdflscape}
\usepackage{subfig}
\usepackage{caption}
\usepackage{amsmath}
\let\And\and

\title{Problems With Large Language Models for Learner Modelling: Why LLMs Alone Fall Short for Responsible Tutoring in K--12 Education}

\author{
 Danial Hooshyar \\
  School of Digital Technologies\\
  Tallinn University, Tallinn\\
  Estonia\\
  Faculty of Information Technology\\
  Faculty of Education and Psychology\\
  University of Jyv\"askyl\"a, Jyv\"askyl\"a\\
  Finland\\
  \texttt{danial.hooshyar@tlu.ee} \\
   \And
 Yeongwook Yang \\
  Department of Computer Science and Engineering\\
  Gangneung-Wonju National University\\
  Republic of Korea \\
  \texttt{yeongwook.yang@gwnu.ac.kr} \\
    \And
 Gustav \v{S}\'{\i}r\\
  Department of Computer Science\\
  Czech Technical University\\
  Czech Republic\\
  \texttt{gustav.sir@cvut.cz} \\
    \And
Tommi K\"arkk\"ainen\\
  Faculty of Information Technology\\
  University of Jyv\"askyl\"a, Jyv\"askyl\"a\\
  Finland\\
  \texttt{tommi.karkkainen@jyu.fi} \\
    \And  
 Raija H\"am\"al\"ainen\\
  Faculty of Education and Psychology\\
  University of Jyv\"askyl\"a, Jyv\"askyl\"a\\
  Finland\\
  \texttt{raija.h.hamalainen@jyu.fi} \\
    \And
 Mutlu Cukurova\\
  UCL Knowledge Lab\\
  UCL Centre for Artificial Intelligence\\
  University College London\\
  UK\\
  \texttt{m.cukurova@ucl.ac.uk} \\
    \And    
 Roger Azevedo \\
  School of Modeling Simulation and Training\\
  University of Central Florida\\
  US\\
  \texttt{roger.azevedo@ucf.edu} \\
}

\date{}
\begin{document}
\maketitle
\begin{abstract}
The rapid rise of large language model (LLM)-based tutors in K--12 education has led to the misconception that generative models can replace traditional learner modelling and act as general-purpose engines for adaptive instruction. This is especially problematic in K--12 settings, which the EU AI Act classifies as a high-risk domain requiring responsible design. Motivated by concerns surrounding the role of learner modelling in responsible AI-powered tutoring, this study synthesises existing research evidence on key limitations of LLM-based tutoring systems and then empirically investigates one critical aspect of these concerns: the accuracy, reliability, and temporal coherence of assessing learners' evolving knowledge over time. To this end, we compare a deep knowledge tracing (DKT) model with a widely used LLM (with and without fine-tuning) that has demonstrated competitive performance in tutoring-related tasks, using a large-scale open-access dataset. Our findings show that DKT achieves the highest discrimination performance (AUC = 0.83) on \textit{next-step correctness prediction} and consistently outperforms the LLM across evaluation settings. Although fine-tuning improves the LLM's AUC by about 8\% over the zero-shot baseline, it still remains 6\% below DKT and produces higher early-sequence errors, precisely where incorrect predictions would be most harmful for adaptive learner support. Temporal-coherence analyses further reveal that while DKT maintains stable, directionally correct mastery updates, LLM variants display substantial temporal weaknesses, including smooth but wrong-direction updates, and, even after fine-tuning, remain inconsistent and unable to match DKT's temporal stability. We also illustrate that these issues persist even though fine-tuned LLMs required nearly 198 hours of continuous high-compute training, far exceeding the computational demands of the lightweight DKT model. Our qualitative analysis of \textit{multi-skill mastery estimation} further shows that, even after fine-tuning, the LLM produced unstable and inconsistent mastery trajectories, whereas DKT maintained smooth and coherent multi-skill updates. Collectively, these findings suggest that LLMs are unlikely to achieve the same positive effect sizes observed in decades long intelligent tutoring systems literature. Responsible LLM use in tutoring systems requires embedding them within hybrid, evidence-based frameworks that incorporate learner modelling to ensure safe, accurate, reliable, and pedagogically sound support. 
\end{abstract}



\section{Introduction}

There is widespread recognition of the role that artificial intelligence (AI) can play in K--12 education. The rapid emergence of generative AI, most prominently large language models (LLMs), has accelerated adoption in schools and research activities \cite{giannakos2025promise}. This rapid uptake has also intensified competition among providers, leading to unprecedented acceleration in development, with over 20 major LLMs released in 2023 alone \cite{setala2025mathematical}. Meanwhile, governments are beginning to institutionalize AI in education \cite{selwyn2025prompting}. For example, Estonia’s AI Leap program mandates AI integration across schools (\url{https://en.tihupe.ee/}). In parallel, major funding bodies such as NordForsk \footnote{\url{https://www.nordforsk.org/calls/call-proposals-responsible-use-artificial-intelligence}} and the EU’s Horizon program \footnote{\url{https://digital-strategy.ec.europa.eu/en/policies/european-ai-research}} increasingly promote responsible AI, with initiatives such as Horizon’s AI innovation package \footnote{\url{https://ec.europa.eu/commission/presscorner/detail/en/ip_24_383}} supporting generative AI innovation and talent development through education and training. At the same time, regulatory frameworks now classify AI in education, alongside healthcare, as a \textbf{high-risk} application \cite{euAIAct2024,saarela2025eu}. This recognition underscores the need for responsible AI, typically defined through principles of fairness, transparency, accountability, and human agency \cite{arrieta2020explainable,eitel2021beyond,hooshyar2025responsibleAI, jakesch2022different,maree2020towards,cerratto2025towards}. While the responsible use of AI in educational settings is widely emphasized in both policy and research, the \textbf{question of how these systems are responsibly developed remains largely overlooked.} 

In practice, the rapid integration of generative AI into educational settings has outpaced efforts to ensure such responsibility at the design/development level. For decades, adaptive educational systems, most notably intelligent tutoring systems (ITSs), have relied on learner modelling approaches grounded in established theories of cognition, pedagogy, and human learning \cite{azevedo2023theories,conati2023student,krivich2025systematic}. These systems maintain explicit learner models which are structured representations of a student’s knowledge, skills, misconceptions, progress, or affective states \cite{abyaa2019learner}. Such representations enable transparent, interpretable, and pedagogically meaningful forms of adaptivity and thus align closely with responsible AI principles. This tradition is now being displaced by the rising use of general-purpose LLMs \cite{hooshyar2025responsibleAI}. Because these models are not developed for educational contexts, they do not incorporate the structured representations or data foundations needed to model learners in principled ways. Instead, the adaptive behaviours they produce stem from opaque, general-purpose training (often for user satisfaction) and exhibit known limitations including hallucinations, reasoning inconsistencies, and opaque decision-making processes \cite{zhao2023survey}. These make LLM behaviours insufficiently reliable in educational contexts. Their appeal stems from convenience: LLMs function as “off-the-shelf” systems built for broad, cross-domain use and can be integrated into classrooms with minimal expertise or development time \cite{giannakos2025promise}. Yet convenience should not be confused with better support for learners. In addition, this convenience conceals a deeper contradiction. Simply labelling tools as “ethical,” “fair,” or “responsible,” or encouraging their responsible use, cannot compensate for the absence of responsible design and development practices. If educational AI is to be truly trustworthy, responsibility should be embedded in its development, not retrofitted in its use.     

\textbf{Responsible use presupposes responsible development}. Technical fixes, such as fine-tuning LLMs to reduce hallucinations or mitigate bias, address surface issues but cannot correct structural flaws. LLMs, trained on statistical patterns rather than conceptual understanding, often conflate factual knowledge with societal bias \cite{lee2024life,resnik2025large}. Hooshyar et al. \cite{hooshyar2025responsibleAI} argue that LLMs may reproduce both neutral knowledge and harmful biases, such as gendered assumptions about success in mathematics, while Du et al. \cite{du2025benchmarking} demonstrate that even state-of-the-art LLMs exhibit systematic and asymmetric gender biases in pedagogical feedback. Recent experimental evidence shows that when relying on LLMs for essay writing, students did not engage the same high levels of neural connectivity, cognitive engagement, metacognitive monitoring, and sense of ownership observed in brain-only and search-based approaches, highlighting measurable declines in learning skills and outcomes \cite{kosmyna2025your,AzevedoWiedbusch2025}. While the study focused on ChatGPT rather than an LLM-based tutor, the findings still underscore important risks for educational use. When it comes to the use of LLMs to support teachers, recent research shows that teachers spend significant time reviewing, repairing, and reworking LLM-based tool outputs to ensure pedagogical appropriateness, underscoring the hidden human labour required to compensate for the educational frailties of such tools \cite{selwyn2025prompting}. The study conducted by Karumbaiah et al. \cite{karumbaiah2024evaluating} showed that behaviour analysis reveals hidden failures of language models in assessing student essays—such as struggles with negation and multilingual writing. 

\begin{figure}[!htbp]
  \centering
  \includegraphics[width=0.9\textwidth]{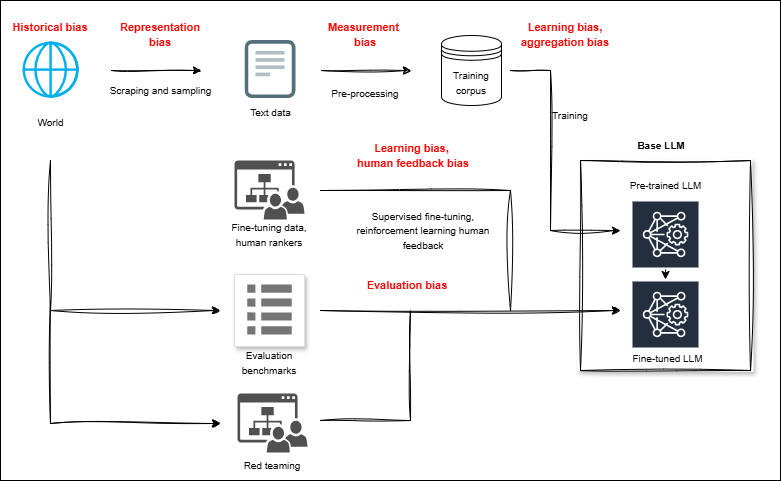}
  \caption{Potential sources of bias during the initial development of LLMs, adapted from Lee et al. \cite{lee2024life}.}
  \label{fig:myfig1}
\end{figure}

Building on these concerns, Fig~\ref{fig:myfig1} illustrates how biases can emerge during the initial development phase of the LLM life cycle. As Lee et al. \cite{lee2024life} explain, biases could enter at multiple stages, from training data collection to evaluation, and are compounded by existing inequities. Web-based corpora such as Common Crawl or WebText \cite{radford2019language} may embed representational imbalances, toxic content, and stereotypes \cite{birhane2023into,selbst2016big}, while pre-processing steps are often constrained by subjective judgments \cite{lee2024life}. During pre-training, LLMs learn patterns from context, but the optimization process can also strengthen stereotypes already present in the data \cite{bolukbasi2016man,goldfarb2021intrinsic}. Post-training alignment through reinforcement learning-based methods, which fine-tune the model’s behaviour to better align with the goals, needs, or preferences of a user group, can also introduce human feedback biases \cite{casper2023open}. Evaluation frameworks such as holistic evaluation of language models \cite{liang2022holistic} include metrics on interpretability, transparency, and toxicity, yet evaluation—whether through benchmarks or adversarial “red teaming,” where external testers deliberately probe models for flaws—remains prone to sampling gaps and selective reporting \cite{young2018model}. \textit{Extending from development to application, fine-tuning or customization does not address certain biases introduced during LLM training and development, and may further introduce risks when base models are tailored for education through supervised fine-tuning, preference tuning, prompt tuning, or retrieval-augmented generation} \cite{lee2024life}. 

Given these limitations of LLMs, carefully developed predictive modelling based on learning data remains essential for education (see section 1.2). Unlike LLMs trained on large, biased web corpora, predictive learner models have long played a central role in learner modelling and the prediction of student performance, dropout, and mastery \cite{conati2023student,hooshyar2024temporal,piech2015deep,krivich2025systematic}. By analysing interaction data, predictive learner models such as Bayesian Knowledge Tracing (BKT) and Deep Knowledge Tracing (DKT) provide interpretable insights into how learners acquire skills, where difficulties arise, and how knowledge evolves over time. \textbf{Recently, however, the powerful ability of LLMs to generate individualized responses has fostered a misconception: that they can replace accurate predictive learner modelling and act as general-purpose engines for AI in education}. This assumption is flawed because LLMs do not construct or maintain a learner model that is an accurate, explicit, data-driven representation of a student’s knowledge states over time. Without such a model, LLMs cannot reliably assess learning or go beyond limited adaptivity, as their judgments are typically too superficial or coarse in comparison to predictive learner models that trace learners’ states over time by monitoring trajectories, estimating mastery, and modelling learning dynamics (e.g., \cite{borchers2025can}). Notably, while LLMs can, to a limited extent, estimate a learner’s likelihood of success on the next task, they lack the capacity to generate stable probabilities of mastery across all skills based solely on the learner’s current sequence of actions, something advanced learner modelling methods like DKT can achieve in real time.  

This paper first examines the principles of responsible AI. It then, through an empirical study comparing DKT and LLM-based tutoring, illustrates one of the methodological challenges that limit LLMs’ capacity to support evidence-based adaptivity in education to make an argument that \textbf{LLMs alone may not be able to serve as responsible tutors in K--12 education}.

\subsection{Defining responsible AI}

The notion of responsible AI has been conceptualized within diverse frameworks, each stressing partially overlapping yet distinct sets of principles. Maree et al. \cite{maree2020towards} emphasize values such as fairness, privacy, accountability, transparency, and soundness. Arrieta et al. \cite{arrieta2020explainable} broaden this perspective by integrating ethics, security, and safety, whereas Eitel-Porter \cite{eitel2021beyond} and Werder et al. \cite{werder2022establishing} identify explainability as a key requirement. Jakesch et al. \cite{jakesch2022different} advance a more comprehensive account, incorporating sustainability, inclusivity, societal benefit, human autonomy, and solidarity. Collectively, these perspectives demonstrate that responsibility in AI extends beyond technical reliability to encompass social, ethical, and human-centred dimensions.

Building on these foundations, this paper adopts the definition proposed by Goellner et al. \cite{goellner2024responsible}, who define responsible AI as \textit{a human-centred approach that builds user trust through ethical and reliable decision-making, explainable outcomes, and privacy-preserving implementation}. When it comes to the design process, one can focus on human-centeredness, ethical and reliable decision-making, and explainability of the AI methods, as these principles can be embedded in model development and, when realized, naturally foster user trust and support privacy-preserving practices \cite{hooshyar2025responsibleAI,hooshyar2025towards}. In the present study, however, we narrow this broad framework to the aspects most foundational for (intelligent) tutoring which is accurate and reliable real-time assessment of learners’ knowledge and progress \cite{d2023intelligent}. A core prerequisite of responsible tutoring is the system’s ability to generate stable, evidence-based adaptive decisions. Thus, rather than addressing explainable or human-centredness, which target additional responsible-AI principles, we focus specifically on the assessment and adaptivity components of responsible tutoring. Our aim is to evaluate whether LLMs demonstrate the fundamental capability to assess and adapt to learners in real time, comparing their performance directly to DKT models. This allows us to determine whether LLMs meet the baseline requirement of responsible (intelligent) tutoring regarding accurate and stable assessment of learners’ evolving states, before engaging with broader responsible AI dimensions. This definition therefore serves as the lens for the present work. \textit{We regard an AI system that does not ensure sound, reliable, and educationally appropriate predictions at the level of development as fundamentally less responsible regardless of whether its later use is framed as “ethical” or “responsible.”}

\subsection{Predictive learner modelling in AI for education}

In the context of K--12 education, AI has primarily been applied for two overarching aims: improving pedagogical practice and supporting research advancement. In practice, classroom-level applications include personalized instruction, adaptive feedback, and scaffolding tailored to learners’ needs \cite{cukurova2020promise,holmes2020aie,holmes2022artificial,hooshyar2022effects,pedro2019artificial,vincent2020trustworthy}. School-level applications extend to dropout prediction and administrative automation \cite{barros2019predictive,nagy2024interpretable}. On the research side, AI is used to extract pedagogical insights such as identifying predictors of math performance from patterns of self-regulated learning behaviour \cite{hooshyar2025towards}.

At the centre of many of these applications lies predictive modelling, most prominently in the form of learner modelling. Learner models construct structured representations of students’ cognitive and non-cognitive characteristics, enabling real-time inference of knowledge states, performance trajectories, and motivational factors \cite{abyaa2019learner,conati2023student,hooshyar2024temporal,krivich2025systematic}. Methodologically, predictive modelling has followed two broad traditions: symbolic and sub-symbolic methods, each with unique advantages and limitations \cite{holmes2022artificial,hooshyar2025responsibleAI}. Symbolic models, such as rule-based systems, offer transparency but struggle with uncertainty, complexity, and scalability \cite{ilkou2020symbolic}. Sub-symbolic methods, such as deep neural networks, often outperform symbolic methods and handle complex (non-)sequential data effectively, but they raise serious concerns about opacity, bias, and lack of pedagogical grounding \cite{garcez2023neurosymbolic,hooshyar2024augmenting,hooshyar2024problems,garcez2019neural}. These concerns are particularly critical in education, where unexplainable or biased systems can reinforce inequality and fail to adapt meaningfully to individual learner needs \cite{baker2022algorithmic}. In response, the field has increasingly turned to explainable AI (XAI) as a step toward more interpretable and trustworthy systems \cite{arrieta2020explainable,hooshyar2022three,saarela2021explainable}. Post-hoc XAI methods such as SHAP and LIME provide explanations of predictions and, to some extent, reveal the reasoning behind the decision-making of AI algorithms. However, multiple studies have recently shown that they often offer only partial or misleading explanations, limiting both interpretability and trust \cite{hooshyar2024problems,lakkaraju2020fool}. More fundamentally, most sub-symbolic models remain fully data-driven, unable to incorporate structured educational knowledge such as causal relationships among variables, theoretical models connecting learning constructs, the role of specific variables in explaining phenomena, or variable interactions affecting learning outcomes which is limiting both pedagogical alignment and interpretability \cite{celik2022promises,hooshyar2025responsibleAI}. The absence of embedded domain knowledge not only increases the risk that predictions (decision-making) are shaped by spurious correlations or hidden data biases \cite{hooshyar2024augmenting,yang2022understanding} but also limits the ability of educators to engage with the models and ensure alignment with pedagogical goals \cite{ayanwale2022teachers,celik2022promises}. 

These methodological tensions place predictive modelling for education at a crossroads: \textit{should the field continue with opaque but high-performing models, or move toward approaches that align predictive accuracy with principles of responsible AI?}

\subsection{The paradigm shift toward general intelligence and LLMs}

\begin{figure}[!htbp]
  \centering
  \includegraphics[width=0.9\textwidth]{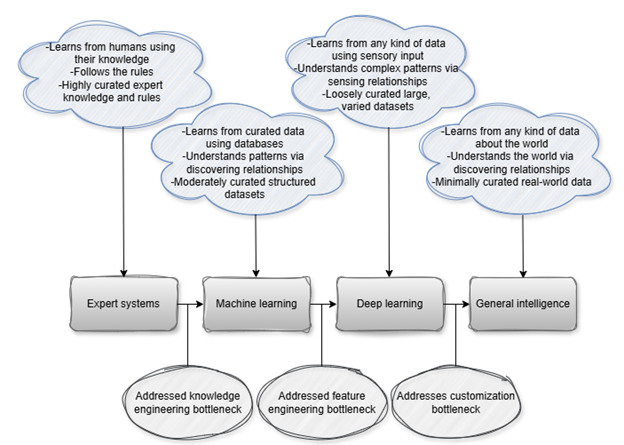}
  \caption{Illustration of the evolving paradigms in artificial intelligence, adapted from Dhar \cite{dhar2024paradigm}.}
  \label{fig:myfig2}
\end{figure}

The evolution of AI reflects successive paradigm shifts. Expert systems depended on curated rules, machine learning on discovering patterns within structured datasets, while deep learning enabled extraction of abstract representations from less curated, real-world data. Each paradigm attempted to address prior limitations, ranging from reliance on knowledge engineering to the need for feature engineering \cite{dhar2024paradigm}. This progression is illustrated in Fig~\ref{fig:myfig2}, which depicts the overall paradigm shift of AI across successive stages. \textbf{Yet rather than addressing the fundamental limitations of sub-symbolic methods (most notably their opacity, bias, and their inability to embed structured domain knowledge), the field turned to training such methods on ever-larger amounts of data} \cite{dhar2024paradigm,miikkulainen2024generative}. The most recent shift toward general intelligence relies on massively pre-trained models trained on unprecedented volumes of text and multimodal data. LLMs, as a form of generative AI, exemplify this trajectory \cite{naveed2025comprehensive}. 

The adoption of LLMs in education has accelerated at an unprecedented pace. These systems have demonstrated clear potential to enhance student engagement in programming \cite{kumar2023impact,lyu2024evaluating}, strengthen writing skills \cite{benvenuti2023artificial}, support teachers through automated tutoring and grading, and deliver on-demand homework assistance to student \cite{labadze2023role,miroyan2025analyzing}. Their fluency and accessibility make them attractive for both classrooms and research, reinforcing their perception as transformative tools. However, these benefits obscure foundational limitations. \textit{LLMs inherit the weaknesses of sub-symbolic deep learning while amplifying them through scale} \cite{lee2024life,yan2024practical}. They remain opaque “black boxes,” with billions of parameters that defy interpretability \cite{hooshyar2025responsibleAI}. As such, the natural-language explanations they produce should not be mistaken for faithful representations of the underlying reasoning. They hallucinate facts, reproduce biases embedded in training data, and lack mechanisms to distinguish between factual knowledge, contingent correlations, and problematic stereotypes, thereby risking confusing students or reinforcing misconceptions \cite{nye2023generative,resnik2025large,werder2022establishing,wang2024large}. Moreover, LLMs are not primarily designed for educational assessment and may struggle to reliably track learner trajectories, estimate skill mastery, or monitor (meta)cognitive and motivational states over time, resulting in limited and often unstable adaptivity \cite{borchers2025can,hooshyar2025responsibleAI}. 

For instance, Borchers and Shou \cite{borchers2025can} showed that while LLMs can generate instructional moves, they exhibit only limited adaptivity compared to learner modelling-based systems like ITSs. In their study, they took 75 real ITS tutoring scenarios, each containing a problem (e.g., 3(2x + 4) – 2 = 16), the student’s correct and incorrect steps, ITS hints, suggested next steps, knowledge components, and even chat history. These were dynamically inserted into prompts so that the LLMs could respond in context. To test adaptivity, they then created five modified versions of each scenario by systematically removing one key element at a time—such as the student’s incorrect steps or the provided hint—resulting in 450 different contexts. Each version was given to three models (Llama3-8B, Llama3-70B, GPT-4o), generating a total of 1,350 responses. The outputs were transformed into text embeddings and compared using randomization tests to see whether the LLMs’ responses actually changed when important context was missing. Results showed limited adaptivity overall, which emerged only when the student’s incorrect steps were removed. Pedagogical soundness was evaluated separately using a validated tutor-training classifier, which judged whether responses praised effort and corrected mistakes indirectly. Here, Llama3-8B achieved higher soundness scores, whereas GPT-4o was strong in instruction-following but tended to give overly direct feedback rather than scaffolding learning. Overall, the findings highlight that current LLMs, unlike ITS, mostly generate feedback based only on surface text rather than adapting to deeper learner data, limiting their potential for supporting student learning.

Beyond these limitations in adaptivity and pedagogical soundness, LLMs also tend to overemphasize automated instruction while neglecting critical aspects of effective education, such as self-regulated learning \cite{delikoura2025superficial,stadler2024cognitive}. Because their development has been driven by scale and fluency rather than by the principles of human-centredness, ethical and explainable decision making, as well as fairness, accountability or pedagogical alignment, they alone cannot be considered as responsible for education. \textbf{LLMs thus represent a major paradigm shift toward general intelligence, but one that amplifies the unresolved issues of sub-symbolic methods}.

\subsection{Toward responsible predictive modelling: Hybrid human–AI intelligence}

As LLM-based generative AI illustrates the unresolved limitations of sub-symbolic dominance, a promising alternative lies in hybrid human–AI intelligence, specifically neural-symbolic AI (NSAI) \cite{garcez2019neural,garcez2023neurosymbolic,hooshyar2021neural}. Through the fusion of symbolic reasoning’s explainability and the computational power of neural networks, NSAI bridges two traditions, long regarded as competing \cite{hooshyar2025responsibleAI,sourek2018lifted}. 

NSAI approaches enable human-centred design by: (i) involving educators and researchers in model development through the direct injection of their symbolic knowledge \cite{hooshyar2024temporal,tato2022infusing}, (ii) mitigating data bias and quality limitations by providing explicit knowledge to compensate for missing or noisy data \cite{ciatto2024symbolic,hooshyar2024augmenting}, (iii) incorporating structured educational knowledge such as causal relationships, theoretical models of learning, the role of specific variables, or interactions that affect learning outcomes—ensuring pedagogical alignment \cite{shakya2021student}, and (iv) enhancing interpretability by making the model’s reasoning and predictions more transparent \cite{besold2021neural,hooshyar2024problems}. These features also advance responsible AI principles, including fostering user trust through explainable, pedagogically sound decisions; supporting ethical decision-making by grounding predictions in educational values rather than opaque correlations; enabling privacy-preserving use by reducing reliance on (sensitive) student data through knowledge augmentation; and reinforcing a human-centred approach by involving educators and learners in system design. This human-in-the-loop framework makes NSAI well-suited to education, strengthening transparency, trust, and ethical responsibility (see Hooshyar et al. \cite{hooshyar2025responsibleAI} for a categorization of neural-symbolic AI paradigms and a detailed explanation of how NSAI can support responsible AI in education).

\section{Do we still need predictive learner modelling, or can LLMs reliably trace learners’ learning states?}

Real-time assessment of learners’ knowledge, skills, and learning states (often referred to as learner modelling) has been extensively investigated in the fields of AI in education, learning analytics, and educational data mining \cite{abdelrahman2023knowledge,abyaa2019learner,conati2023student}. Two of the most established approaches are BKT and DKT \cite{mao2018deep}. These methods employ, respectively, probabilistic graphical models and sequential deep neural networks to process students’ interaction histories with tasks and learning materials. Through this processing, they learn latent representations of learners’ knowledge, detect sequential patterns, and estimate the mastery of skills or concepts \cite{kaser2017dynamic}. Importantly, this paradigm is grounded in learning sciences: much like a teacher observing students’ actions and responses to tasks and inferring their understanding in real time, learner modelling systems aim to infer knowledge states from digital interaction traces, thereby supporting adaptive teaching strategies \cite{d2023intelligent}. In short, learner modelling captures the learner’s evolving knowledge—what has been mastered, where difficulties emerge, and how understanding develops over time. Recent work enhances this paradigm by integrating pre-trained language models (e.g., BERT) to generate richer semantic representations of exercises and student responses \cite{lee2024difficulty,tan2021bidkt}. At the same time, context-aware approaches show that leveraging textual and conceptual features of questions (such as difficulty and semantic relations) can mitigate cold-start issues and improve knowledge-tracing accuracy \cite{pandey2020rkt,su2018exercise}. 

In contrast, LLMs operate according to a fundamentally different logic. While they can generate adaptive responses to learners’ prompts and actions, LLMs are not trained on sequences of student interaction data that capture learning dynamics \cite{wang2025llm}. Instead, they are pretrained on vast corpora of general textual data, designed primarily to approximate general-purpose intelligence through next-token prediction \cite{kapoor2024large,resnik2025large}. This distinction has important consequences. Whereas knowledge tracing methods explicitly encode temporal dependencies such as practice effects, forgetting, and transfer \cite{krivich2025systematic}, LLMs lack mechanisms for modelling learners’ evolving states across time. Notably, while LLMs can, to a limited extent, estimate a learner’s likelihood of success on the next task, they lack the capacity to generate stable probabilities of mastery across all skills based solely on the learner’s current sequence of actions. Moreover, although LLMs excel at modelling sequential dependencies in text, they struggle to capture student interaction patterns when exercises or skill IDs are represented as arbitrary numeric tokens without meaningful embeddings, causing the semantic context of learning activities to be lost. Even with fine-tuning, such limitations persist and restrict the model’s ability to represent learners’ evolving knowledge accurately. Similarly, LLMs face difficulties in accurately modelling long problem-solving histories, where hundreds of questions (each with rich textual features) must be considered to trace learning trajectories \cite{wang2025llm}. As a result, while LLMs can respond contextually in the moment, they do not build persistent or reliable models of an individual learner’s knowledge trajectory \cite{kapoor2024large,resnik2025large,wang2025llm}. 

Generative AI methods embed textual or multimodal learner inputs into high-dimensional vector spaces, allowing them to retrieve semantically similar patterns from their training distribution and generate contextually relevant responses \cite{riley2024education}. This enables them to answer questions, provide explanations, and scaffold problem-solving in highly flexible ways. However, unlike learner modelling approaches, they cannot reliably incorporate structured sequences of learner-system interactions into models of skill acquisition \cite{cho2024systematic,park2025comprehensive,riley2024education,wang2025llm}. Moreover, even when LLMs provide answers with apparent confidence, they are often poorly calibrated in representing uncertainty, meaning that their outputs can be misleading or even risky to act upon in educational decision-making \cite{cho2024systematic,kapoor2024large}. 

Furthermore, interpretability is a central concern when considering the role of AI in education. Learner modelling methods, whether Bayesian or neural, are designed to make their reasoning at least partially transparent \cite{bai2024survey,hooshyar2024temporal, krivich2025systematic, yeung2019deep}. For example, they can trace how particular interaction data (e.g., sequences of correct or incorrect responses, number of attempts, or time-on-task) lead to inferences about mastery, forgetting, or misconceptions. Their outputs can often be linked back to specific model components, such as transition probabilities in BKT or weight updates in DKT, thus providing a traceable path from data to prediction. This transparency allows researchers and educators to understand why a prediction was made, to inspect which data points influenced the outcome, and to align the model’s reasoning with established theories of learning \cite{hooshyar2024augmenting,conati2018ai}. By contrast, LLMs are trained on vast, heterogeneous corpora, where the sources of data and their influence on specific outputs are largely opaque. Their predictions emerge from distributed statistical associations across billions of parameters, without a clear mapping between input, internal reasoning, and output \cite{resnik2025large}. As such, they cannot explain why a particular response was generated, nor can they provide an interpretable account of the decision path or underlying assumptions \cite{hooshyar2025responsibleAI,singh2024rethinking}. This opacity severely limits their use in contexts where accountability and trustworthiness are paramount—such as tracing individual learners’ states or making adaptive instructional decisions.

\section{Deep knowledge tracing vs LLM: An empirical study}

A core principle of responsible (intelligent) tutoring lies in its capacity for accurate, real-time assessment of learners’ knowledge and progress, an ability that enables stable, accurate, and evidence-based adaptivity to deal with learner’s real-time individual needs. While responsible AI in education also encompasses other dimensions (e.g., human-centred, transparent, and ethical decision-making), \textit{real-time learner assessment remains a primary prerequisite for any system claiming to be adaptive}. In this study, we therefore focus specifically on the assessment and adaptivity components of responsible tutoring. To do so, we compare LLM models with DKT. This focus allows us to evaluate whether LLMs satisfy the basic prerequisites of responsible tutoring regarding the provision of accurate, stable estimates of learners’ evolving knowledge and adapting accordingly.

\subsection{Dataset and study context}

We use the ASSISTments 2009-2010 non skill-builder dataset, comprising 603,287 logged student–item interactions from the ASSISTments online tutoring platform \cite{feng2009addressing}. Each record corresponds to a student attempt and includes detailed metadata such as problem identifiers, correctness, attempt and hint counts, response time, and associated skill tags (e.g., order\_id, user\_id, problem\_id, correct, skill\_id, skill\_name, hint\_count, opportunity). ASSISTments is a long-running digital platform used in K--12 mathematics, widely studied in AI in Education and knowledge tracing research; its public data releases standardize identifiers and log fields, enabling reproducible modelling across studies. The non skill-builder dataset differs from the skill builder version in that items may be associated with multiple skills, represented as comma-separated lists within the same row rather than duplicated across entries. In this study, we retained only single-skill interactions to simplify sequence construction and maintain consistency with conventional DKT formulations. This design choice ensures a clear skill-to-problem mapping and does not affect the comparison between DKT and LLM-based assessment, as both approaches process the same underlying interaction stream. 

After removing missing entries, we sorted all interactions by student and temporal order (order\_id), constructing chronological sequences of learner activity for each student. Each sequence encodes ordered triplets of skill ($s$), problem ($q$), and correctness ($y$), that is, $(s,q,y)$, representing the learner’s evolving performance. Students with fewer than three recorded interactions were excluded. The resulting dataset contains time-ordered, single-skill sequences, suitable for training sequential learner modelling frameworks such as DKT and for comparing trajectory-based mastery estimation with LLM-based adaptive responses. Thereafter, we randomly split the data by student into training (80\%), validation (10\%), and test (10\%) sets. Splitting by student, rather than by interaction, prevents information leakage across subsets and enables a fair evaluation of the model’s generalization to unseen learners. Table ~\ref{tab:dataset_stats} reports summary statistics for the dataset at four stages: the original log, the dataset after pre-processing, and the train/validation/test splits.  

\begin{table}[!htb]
\centering
\small
\caption{Summary statistics of the dataset.}
\label{tab:dataset_stats}
\begin{tabular}{lccccc}
\hline
\textbf{Statistic} & \textbf{Original} & \textbf{After pre-processing} & \textbf{Train} & \textbf{Val} & \textbf{Test} \\
\hline
\# of records (interactions) & 603128 & 450146 & 321339 & 43602 & 43470 \\
\# of students              & 8096   & 7981   & 5784   & 723   & 724   \\
\# of quizzes               & 6907   & 6030   & 5338   & 3701  & 3368  \\
\# of skills                & 200    & 149    & 147    & 145   & 142   \\
Avg. interactions per student & 74.5 & 56.4 & 55.6 & 60.3 & 60.0 \\
Avg. interactions per quiz    & 87.3 & 74.7 & 60.2 & 11.8 & 12.9 \\
Avg. interactions per skill   & 3015.6 & 3021.1 & 2186.0 & 300.7 & 306.1 \\
Correct interactions ($y=1$)  & 344680 & 260902 & 186917 & 25250 & 25506 \\
Incorrect interactions ($y=0$)& 258448 & 189244 & 134422 & 18352 & 17964 \\
\hline
\end{tabular}
\end{table}

Fig ~\ref{fig:myfig3} illustrates the characteristics of the processed dataset before splitting. Fig~\ref{fig:myfig3}a shows a highly skewed distribution of interactions per student. Most students attempted fewer than 50 problems, while a small number completed several hundred. Fig~\ref{fig:myfig3}b and ~\ref{fig:myfig3}c present the top 20 most frequent quizzes and skills, respectively, indicating that a small subset of items and mathematical skills accounts for the majority of activity.

\begin{figure}[!htbp]
  \centering
  \includegraphics[width=0.8\textwidth]{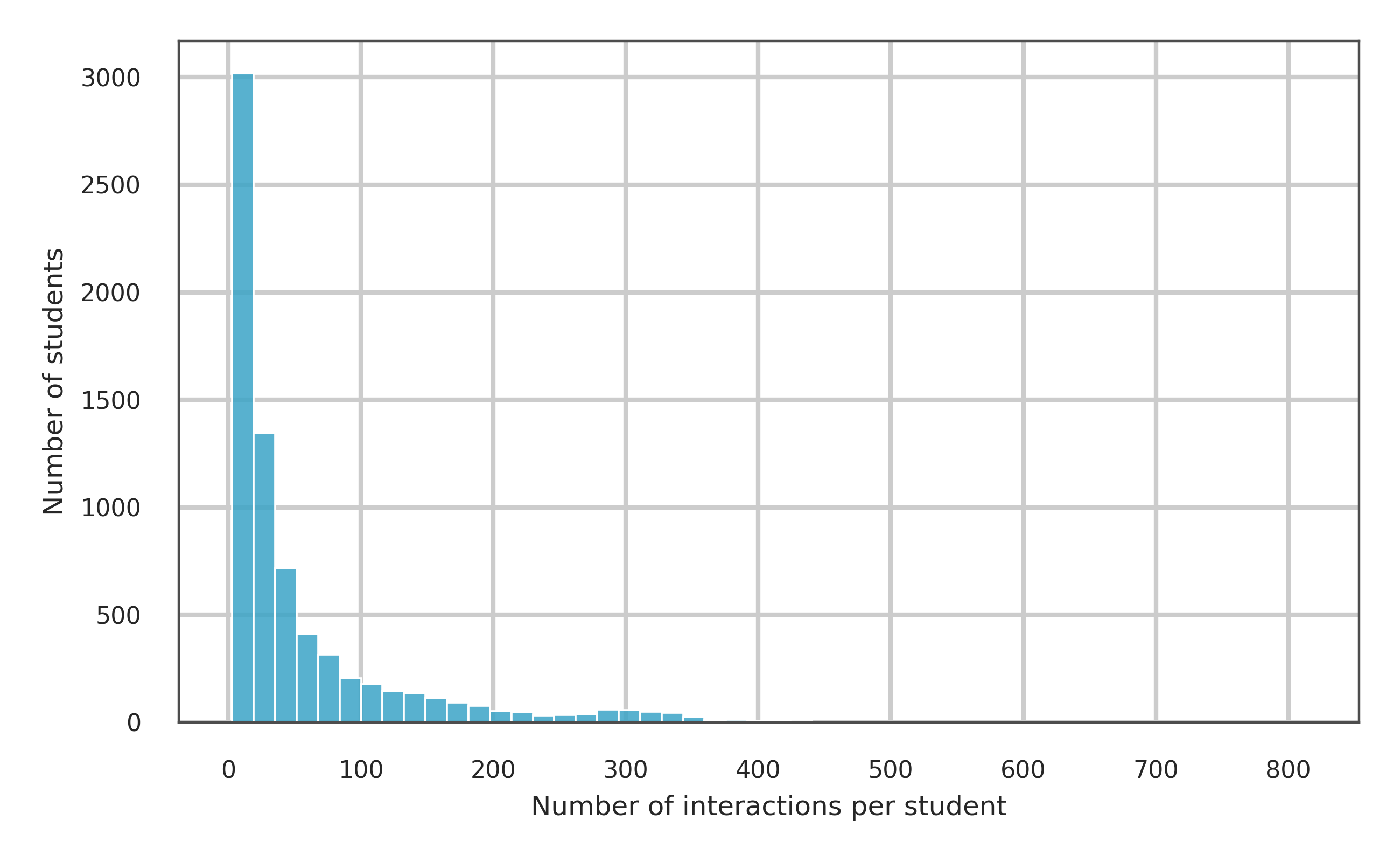}

  \vspace{0.5em}
  \normalsize (a)

  \vspace{1em}
  \includegraphics[width=0.8\textwidth]{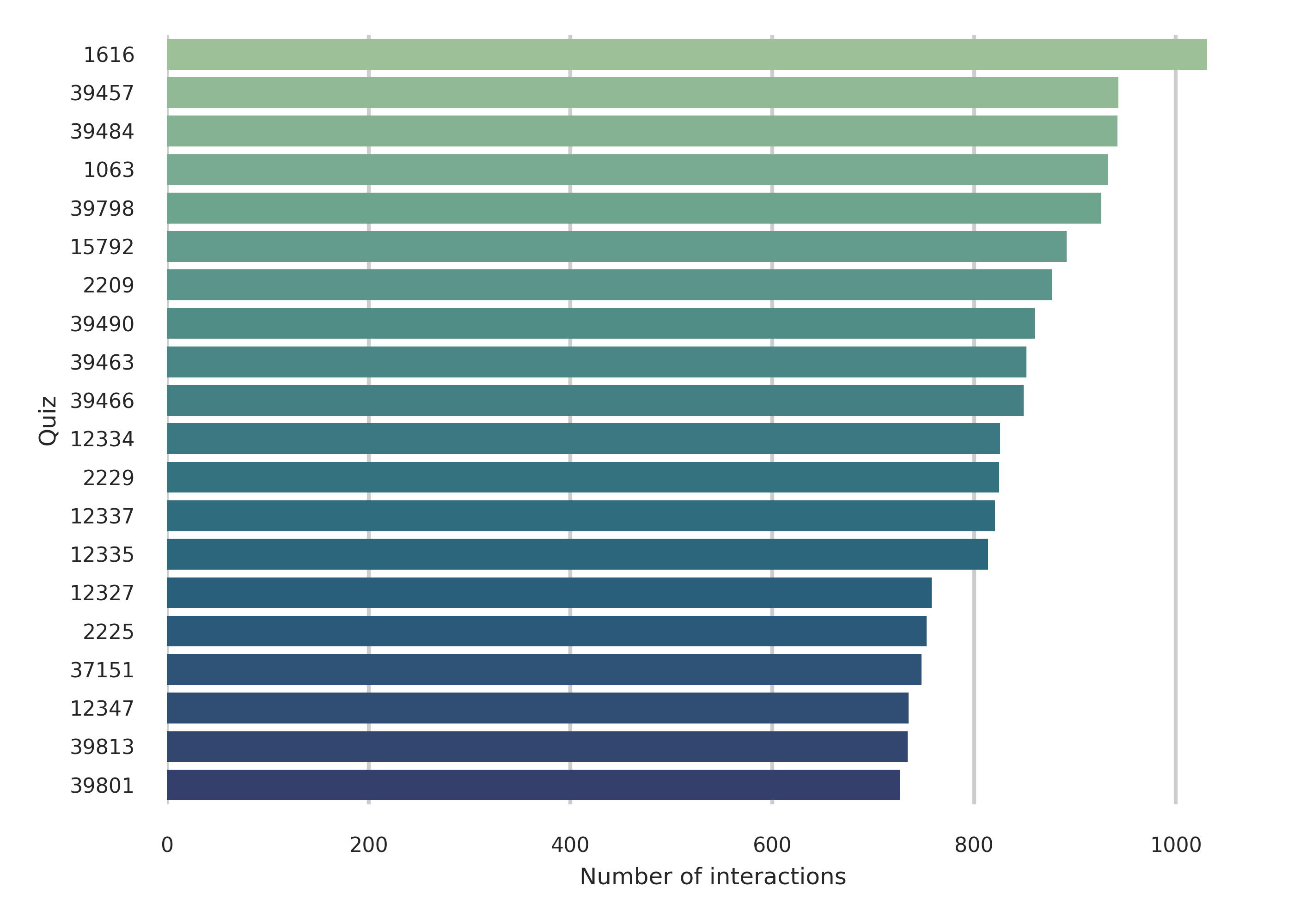}

  \vspace{0.5em}
  \normalsize (b)
    \phantomcaption
\end{figure}

\begin{figure}[!htbp]\ContinuedFloat
  \centering
  \includegraphics[width=0.8\textwidth]{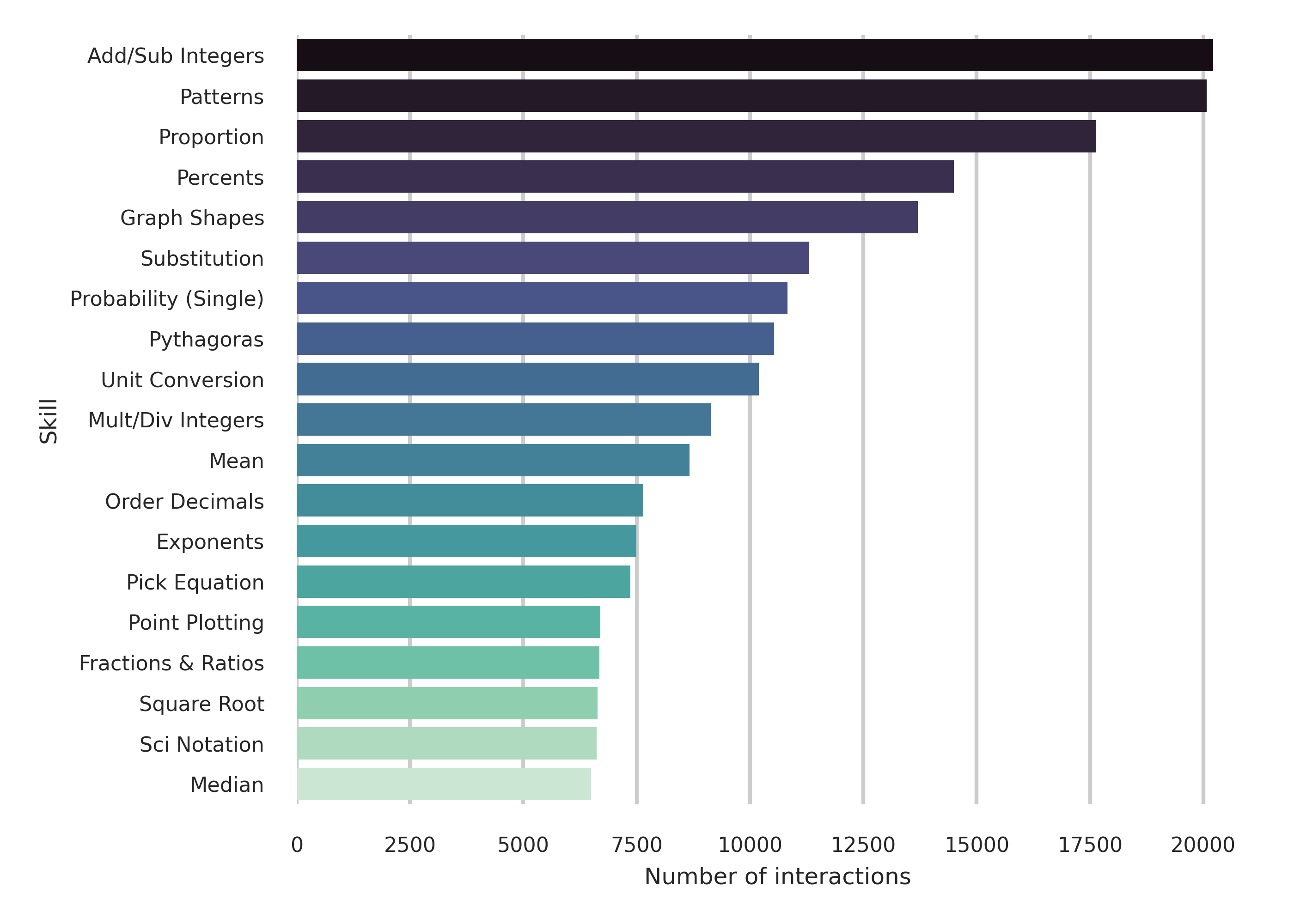}

  \vspace{0.5em}
  \normalsize (c)

  \caption{Dataset characteristics after pre-processing: (a) distribution of interactions per student, (b) top 20 most frequent quizzes, and (c) top 20 most frequent skills.}
  \label{fig:myfig3}
\end{figure}

\subsection{DKT model and LLMs}
\subsubsection{DKT model development}

The DKT model was implemented using a recurrent neural network (RNN) in PyTorch. Each student's learning history was represented as an ordered sequence of triplets (\textit{skill}, \textit{quiz}, \textit{correctness}), describing the practiced skill, the attempted item, and the observed outcome. Each interaction was encoded using the standard $2K$ DKT encoding, where each skill is represented twice, once for incorrect responses and once for correct responses. Formally, for a dataset with $K$ skills, each item was mapped to an integer index

\begin{equation}
x_t = s_t + y_t \cdot K,
\end{equation}

producing a single input sequence $X$ of length $T$. The model also retained the parallel sequences of skill identifiers $S$ and correctness labels $Y$, because $x_t$ encodes what happened at time step $t$, whereas the raw skill ID $s_{t+1}$ is needed to select the correct output component for the next-step prediction target. This allows the model to extract $y_{t+1}$ for the appropriate skill during training. To ensure that loss was computed only on meaningful prediction positions, the model constructed a mask matrix $W$ for each batch:

\begin{equation}
W_{i,t} =
\begin{cases}
1, & \text{if timestep } t \text{ has a valid next-step target}, \\
0, & \text{otherwise}.
\end{cases}
\label{eq:mask}
\end{equation}

This mask removed padded positions and the final timestep from the loss computation, aligning precisely with the DKT next-step prediction objective. 

The architecture consisted of a 64-dimensional embedding layer applied to the $2K$  indices, followed by a single-layer GRU with 128 hidden units. A linear output layer with a sigmoid activation produced a probability over all skills at each timestep. The model’s output at time $t$ was used to predict the correctness label for time $t+1$:
\begin{equation}
\hat{y}_{t+1} = p_t(s_{t+1}),
\end{equation}

extracted by selecting the probability corresponding to the next skill.

Binary cross-entropy loss was minimized using the Adam optimizer (learning rate = 0.001), with gradient clipping and early stopping (patience = 3) based on validation loss. The overall steps of the DKT training procedure are summarised below:

\begin{enumerate}
    \item Construct per-student sequences of (\textit{skill}, \textit{quiz}, \textit{correctness}).
    \item Build aligned sequences $\mathbf{X}$ ($2K$ encoding), $\mathbf{S}$ (skills), and $\mathbf{Y}$ (labels).
    \item Pad variable-length sequences and pack them for efficient RNN processing.
    \item Compute mastery probabilities for all skills at each timestep.
    \item Shift the $\mathbf{S}$ and $\mathbf{Y}$ sequences to form next-step prediction targets.
    \item Build the mask matrix $\mathbf{W}$ to identify valid prediction positions.
    \item Apply the mask and compute binary cross-entropy loss only on valid positions.
    \item Update parameters using Adam; apply gradient clipping and early stopping.
\end{enumerate}

At inference, the model outputs a vector of predicted mastery probabilities for all skills after each student interaction, providing a dynamic representation of evolving knowledge. These continuous mastery trajectories are later compared with adaptive responses generated by LLM models. 

\subsubsection{LLM models: Zero-shot baseline and fine-tuned variant}

For comparison, an LLM (Llama 3 8B: Dubey et al. \cite{dubey2024llama}) was deployed locally using Ollama\footnote{https://ollama.com/library/llama3.1:8b}  to replicate the DKT testing setup. This model was selected because prior AI in education research has shown that it demonstrates comparatively strong adaptive behaviour relative to newer and larger models in tutoring-related tasks (e.g., \cite{borchers2025can,du2025benchmarking,lv2025genal}). Because verbalized numeric probabilities would introduce token-level artifacts and calibration inconsistencies, the LLM was constrained to output only the tokens “0” or “1”, and all correctness probabilities were derived directly from the model’s logits for these two output tokens. This ensures methodological consistency and avoids relying on textual numeric interpolation. Inference was performed deterministically (temperature = 0), and all experiments were repeated twice to verify output stability. We also fine-tuned the same Llama 3 8B model using the training portion of our dataset. The model was updated using a parameter-efficient LoRA configuration to avoid full-model weight updates \cite{hu2022lora}. Fine-tuning was performed for a maximum of 10 epochs with a batch size of 16 and a gradient-accumulation factor of 2. The LoRA hyperparameters were set to rank = 32, $\alpha = 32$, and dropout = 0.1. Training used the Adam optimizer with a learning rate of $3 \times 10^{-4}$, weight decay of $1 \times 10^{-5}$, and a cosine learning-rate scheduler. Early stopping based on validation loss was employed to prevent overfitting. During fine-tuning, each input consisted of a student’s historical interaction sequence (quiz ID, skill name, and correctness), and the model was trained to output only the tokens “0” or “1” for next-step correctness. At inference time, we derived probabilities directly from the logits associated with these two tokens. This setup enabled a direct comparison between the base LLM, the fine-tuned LLM, and the DKT model under identical data splits and evaluation conditions. 

Table ~\ref{tab:prompt_example} presents the main prompt format used in our experiments. Both the zero-shot and fine-tuned models received all previous learner interactions as context and produced a binary output (0 or 1). In all cases, probabilities were computed internally from logits. This setup provides a clean and controlled comparison to the DKT model, which also outputs correctness probabilities for each time step. For skill-wise mastery estimation, the model similarly generated one binary prediction per skill, and the same probability-conversion procedure was applied. This yielded a full vector of mastery probabilities at each time step.

\begin{table}[!htb]
\centering
\small
\caption{Example prompt for next-step correctness prediction.}
\label{tab:prompt_example}
\begin{tabular}{p{0.22\linewidth} p{0.73\linewidth}}
\toprule
\textbf{Component} & \textbf{Prompt content} \\
\midrule
\textbf{System message} &
``You are a classification model. Output only a single token: either 0 or 1. Do not generate explanations or additional text.'' \\
\addlinespace
\textbf{User message} &
The following is a student's problem-solving history. Predict whether the next answer will be correct (1) or incorrect (0).

Student's past performance:
\begin{enumerate}
  \item Quiz 3948 (Skill: Addition and Subtraction Integers) $\rightarrow$ Correct
  \item Quiz 3949 (Skill: Addition and Subtraction Integers) $\rightarrow$ Incorrect
\end{enumerate}

Next quiz: Quiz 3950 (Skill: Addition and Subtraction Integers) \\
\bottomrule
\end{tabular}
\end{table}

\subsection{Model evaluation}

Models were evaluated on a held-out test set using next-step prediction, where each record corresponds to a student–quiz pair with a ground-truth label and a 0/1 prediction from both the DKT and LLM models. Model performance was assessed using standard classification metrics, with the Area Under the ROC Curve (AUC) serving as the primary measure of discrimination, complemented by accuracy, precision, recall, and F1-scores derived from confusion matrices at a 0.5 threshold. Receiver Operating Characteristic (ROC) curves were further plotted to visualize the trade-off between true and false positive rates for both models on the common evaluation subset.

To provide a more granular understanding of how model behaviour evolves across learning sequences, additional analyses were conducted beyond global metrics. First, an optimal decision threshold for each model was estimated using Youden’s $J$ statistic from the ROC curve, enabling threshold-adjusted comparisons of early, mid, and late sequence prediction errors. Second, students were grouped into \textit{stable} versus \textit{switching} behavioural profiles based on the number of changes in their ground-truth correctness sequences. A switch was defined as any point where correctness moved from correct to incorrect ($1 \rightarrow 0$) or incorrect to correct ($0 \rightarrow 1$). Students with fewer than two switches (e.g., \textit{1, 1, 1, 0, 0}) were classified as stable, whereas those with two or more switches (e.g., \textit{1, 0, 1, 0, 1}) were classified as switching. For each group, early, mid, and late sequence errors were computed as the proportion of mismatches between predicted and actual responses (e.g., if 7 out of 10 predictions were correct, the error $= 1 - 0.7 = 0.30$). This analysis allowed us to examine whether the models were more effective for learners who progressed steadily or for those exhibiting irregular or unpredictable performance patterns.

Temporal coherence of the predicted mastery curves was additionally assessed using smoothness-related metrics that quantify how consistently each model updates its mastery estimates over time. Specifically, \textit{volatility} was computed as the average absolute change in predicted mastery between consecutive attempts on the same skill, reflecting how stable or jumpy the probability trajectories are:

\begin{equation}
\frac{1}{T - 1} \sum_{t=2}^{T} \left| P_t - P_{t-1} \right|
\label{eq:volatility}
\end{equation}

In parallel, an \textit{inconsistency} rate was calculated to capture how often the model updated mastery in a direction that contradicted the student’s observed response (for example, decreasing mastery after a correct answer or increasing it after an incorrect one). Formally, it was computed as the proportion of sign mismatches between 

\begin{equation}
\Delta P_t = P_t - P_{t-1}
\label{eq:deltaP}
\end{equation}

and the expected direction determined by the ground-truth response. Together, these two measures provide insight into the stability and pedagogical correctness of each model’s temporal learning behaviour. Finally, sequential stability of predictions was analysed using \textit{multi-skill mastery heatmaps} generated from the LLM and the DKT model’s full-skill predictions. These visualisations compared both models’ probability trajectories across time for a selected student, enabling qualitative assessment of \textit{consistency, smoothness, and responsiveness} in multi-skill prediction dynamics. Together, this multi-level evaluation framework provided both broad and fine-grained comparisons between the neural-sequential (DKT) and language-based (LLM) models in learner modelling and accordingly predicting learner performance. In addition to accuracy-focused metrics, \textit{computational efficiency} was assessed to contextualize each model’s practicality for real-world deployment, including the time required for training and inference as well as the hardware resources needed to produce predictions.

\section{Results and analysis}
\subsection{Quantitative analysis}

\begin{figure}[!htbp]
  \centering
  \includegraphics[width=0.7\textwidth]{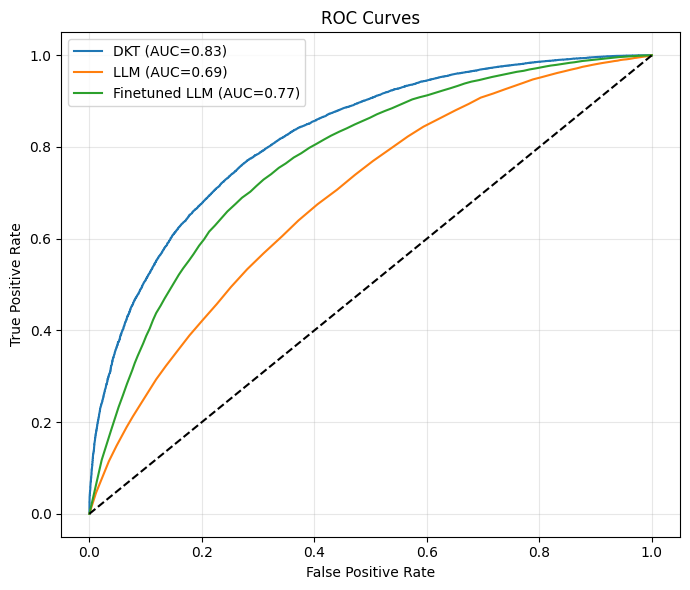}
  \caption{ROC curves comparing prediction performance across all models.}
  \label{fig:myfig4}
\end{figure}

Table ~\ref{tab:model_performance} and Fig ~\ref{fig:myfig4} together show clear performance differences between the DKT model and the LLM-based approaches. Across all metrics, DKT remains the strongest overall performer, achieving the highest AUC (0.83) and notably higher accuracy, precision, recall, and F1-scores for both low and high performers. In contrast, the zero-shot Llama 3 show substantially weaker predictive power, with AUC score of 0.69. This instability is further visible in the wider variation across F1-score patterns, particularly for low-performing students. The fine-tuned Llama 3 model provides a significant improvement over zero-shot baseline, raising the AUC to 0.77 and closing part of the performance gap with DKT. However, even with fine-tuning, the LLM still does not match the stable predictive behaviour of DKT, which delivers consistently higher discrimination ability and more reliable classification metrics.

\begin{table}[!htb]
\centering
\small
\caption{Performance of baseline DKT model and LLM variants.}
\label{tab:model_performance}
\begin{tabular}{llcc}
\hline
\textbf{Model} & \textbf{Metric} & \textbf{Low Performer} & \textbf{High Performer} \\
\hline
\multirow{5}{*}{DKT}
 & AUC (\%)        & \multicolumn{2}{c}{83} \\
 & Accuracy (\%)   & \multicolumn{2}{c}{75} \\
 & Precision (\%)  & 73 & 77 \\
 & Recall (\%)     & 64 & 83 \\
 & F1-score (\%)   & 68 & 80 \\
\hline
\multirow{5}{*}{Llama 3 8B (zero-shot)}
 & AUC (\%)        & \multicolumn{2}{c}{69} \\
 & Accuracy (\%)   & \multicolumn{2}{c}{64} \\
 & Precision (\%)  & 74 & 63 \\
 & Recall (\%)     & 20 & 95 \\
 & F1-score (\%)   & 31 & 76 \\
\hline
\multirow{5}{*}{Llama 3 8B (fine-tuned)}
 & AUC (\%)        & \multicolumn{2}{c}{77} \\
 & Accuracy (\%)   & \multicolumn{2}{c}{72} \\
 & Precision (\%)  & 67 & 75 \\
 & Recall (\%)     & 64 & 78 \\
 & F1-score (\%)   & 65 & 76 \\
\hline
\end{tabular}
\end{table}

Table ~\ref{tab:threshold_analysis} presents how model behaviour evolves across learning sequences by summarising early-, mid-, and late-stage errors for both stable and switching students, using each model’s own optimal ROC threshold. Across all models, error rates were consistently lower for stable students than for switching students, reflecting the greater difficulty of predicting learners whose performance fluctuates. Among the models, DKT achieved the lowest errors overall and most notably in the early stage (0.3118 for switching; 0.2975 for stable students). This is important because early misclassifications can delay appropriate instructional support or trigger unnecessary interventions, potentially hindering learning before students gain momentum. DKT also maintained the strongest performance in the middle and late sequence segments, indicating consistent reliability as students progress. While the baseline LLM showed higher errors (particularly for switching students), they still followed the expected pattern of decreasing errors over time. The fine-tuned LLM, however, demonstrated improvements over its base forms, narrowing the performance gap and showing that domain-specific adaptation enhances prediction stability.

\begin{table}[!htb]
\centering
\small
\caption{Performance of baseline DKT model and LLMs.}
\label{tab:threshold_analysis}
\begin{tabular}{lccccccc}
\hline
\textbf{Model} & \textbf{Threshold*} & \multicolumn{3}{c}{\textbf{Switching}} & \multicolumn{3}{c}{\textbf{Stable}} \\
\cline{3-5}\cline{6-8}
 &  & \textbf{Early} & \textbf{Middle} & \textbf{Late} & \textbf{Early} & \textbf{Middle} & \textbf{Late} \\
\hline
DKT            & 0.5782 & 0.3118 & 0.2964 & 0.2742 & 0.2975 & 0.1469 & 0.1217 \\
LLM            & 0.8789 & 0.4054 & 0.4018 & 0.3841 & 0.3853 & 0.2886 & 0.2487 \\
Fine-tuned LLM & 0.6211 & 0.3309 & 0.3251 & 0.3430 & 0.3563 & 0.1803 & 0.2501 \\
\hline
\end{tabular}

\vspace{0.5ex}
\footnotesize{*Optimal ROC threshold.}
\end{table}

As shown in Table ~\ref{tab:temporal_coherence}, the temporal coherence metrics reveal important differences in how each model updates its mastery estimates over time. DKT shows the most balanced and pedagogically appropriate behaviour, with the lowest volatility (0.1075) and inconsistency among all models (0.4061). This indicates that its probability updates are generally stable and move in the correct direction relative to the learner’s performance. By contrast, the baseline LLM variant exhibits clear temporal weaknesses, showing higher inconsistency (0.5012), meaning that many of its updates contradict what the student’s response suggests. Although its volatility is relatively low (0.1157), this apparent stability is not accompanied by pedagogically meaningful behaviour, as its updates often move in the wrong direction, for instance, increasing mastery after an incorrect answer or decreasing it after a correct one.
The fine-tuned LLM shows clear improvements over the zero-shot variants because exposure to real student data teaches it how mastery should rise after correct responses and fall after incorrect ones. This training reduces its inconsistency rate (0.4525) relative to the zero-shot models and produces stronger, more meaningful directional updates. However, this comes with higher volatility, indicating larger probability swings. \textit{While fine-tuning moves the LLM closer to pedagogically appropriate behaviour, it still does not reach the temporal stability of DKT, which remains the most consistent and coherent overall}.

\begin{table}[!htb]
\centering
\small
\caption{Temporal coherence metrics.}
\label{tab:temporal_coherence}
\begin{tabular}{lcc}
\hline
\textbf{Model} & \textbf{Volatility} & \textbf{Inconsistency} \\
\hline
DKT           & 0.1075 & 0.4061 \\
LLM           & 0.1157 & 0.5012 \\
Finetuned LLM& 0.2945 & 0.4525 \\
\hline
\end{tabular}
\end{table}

Table ~\ref{tab:computational_efficiency} illustrates the computational requirements for each evaluated model. The results show that DKT trained and generated predictions within seconds on modest hardware and was even able to train on a standard Google Colab CPU with 12.7 GB of memory in approximately 18 minutes. In contrast, the zero-shot Llama 3 model required 1623.26 seconds (0.4509 hours) for inference, while the finetuned Llama 3 model demanded even greater computation, including 197.969 hours of training time on high-end GPUs. This contrast highlights the substantial computational and resource demands of LLM-based learner modelling compared with the lightweight and highly practical DKT approach.

\begin{table}[!htb]
\centering
\small
\caption{Computational efficiency of the models.}
\label{tab:computational_efficiency}
\begin{tabular}{lccc}
\hline
\textbf{Model} & \multicolumn{2}{c}{\textbf{Computation time}} & \textbf{Resource} \\
\cline{2-3}
 & \textbf{Training time} & \textbf{Testing time*} &  \\
\hline
DKT
 & 50.778 s
 & 31.272 s (0.008687 h)
 & Colab T4 GPU (16GB), 12.7GB RAM \\
LLama 3 8B (zero-shot)
 & --
 & 1623.26 s (0.45 h)
 & Dual NVIDIA H100 (80GB), 320GB RAM \\
LLama 3 8B (fine-tuned)
 & 712,692 s (197.969 h)
 & 1764.68 s (0.49 h)
 & Dual NVIDIA H100 (80GB), 320GB RAM \\
\hline
\end{tabular}

\vspace{0.5ex}
\footnotesize{*Testing time refers to the total inference time required by models to generate predicted probabilities for all instances in the test set.}
\end{table}

\subsection{Qualitative analysis}

Building on the quantitative analysis, which showed that both the DKT model and the fine-tuned LLM outperform the zero-shot LLM across multiple evaluation metrics, this section provides a qualitative comparison of how the models behave over time. When comparing the multi-skill mastery heatmaps generated by the DKT model and the two LLM-based models, clear differences emerge in how each model captures temporal stability and relational structure in student learning. Fig~\ref{fig:myfig5}a and ~\ref{fig:myfig5}b illustrate the predicted skill-mastery probabilities for the test-set student with ID 81439, produced by the DKT model and the fine-tuned LLM, respectively. The DKT model produces smooth, gradual transitions in mastery probabilities across consecutive time steps, reflecting its ability to maintain a coherent temporal memory of learning progress. The fine-tuned LLM still exhibits sharper updates and more fluctuation than DKT, indicating a more local, step-wise pattern of prediction. This is visible in the heatmaps, where highlighted mistakes (shown through annotated cell probabilities) contrast with the white reference lines indicating correct, temporally consistent predictions, corresponding to the overall error counts (DKT: 9; fine-tuned LLM: 19). This contrast underscores how specialised learner modelling approaches with explicit temporal representations, particularly those trained directly on learning process data, are better able to capture the gradual evolution of learner knowledge, producing mastery trajectories that more faithfully reflect real learning processes.

\clearpage
\begin{landscape}
\thispagestyle{empty}

\begin{figure}[p]
  \centering
  \vspace*{-2.2cm} 
  \includegraphics[width=0.85\linewidth]{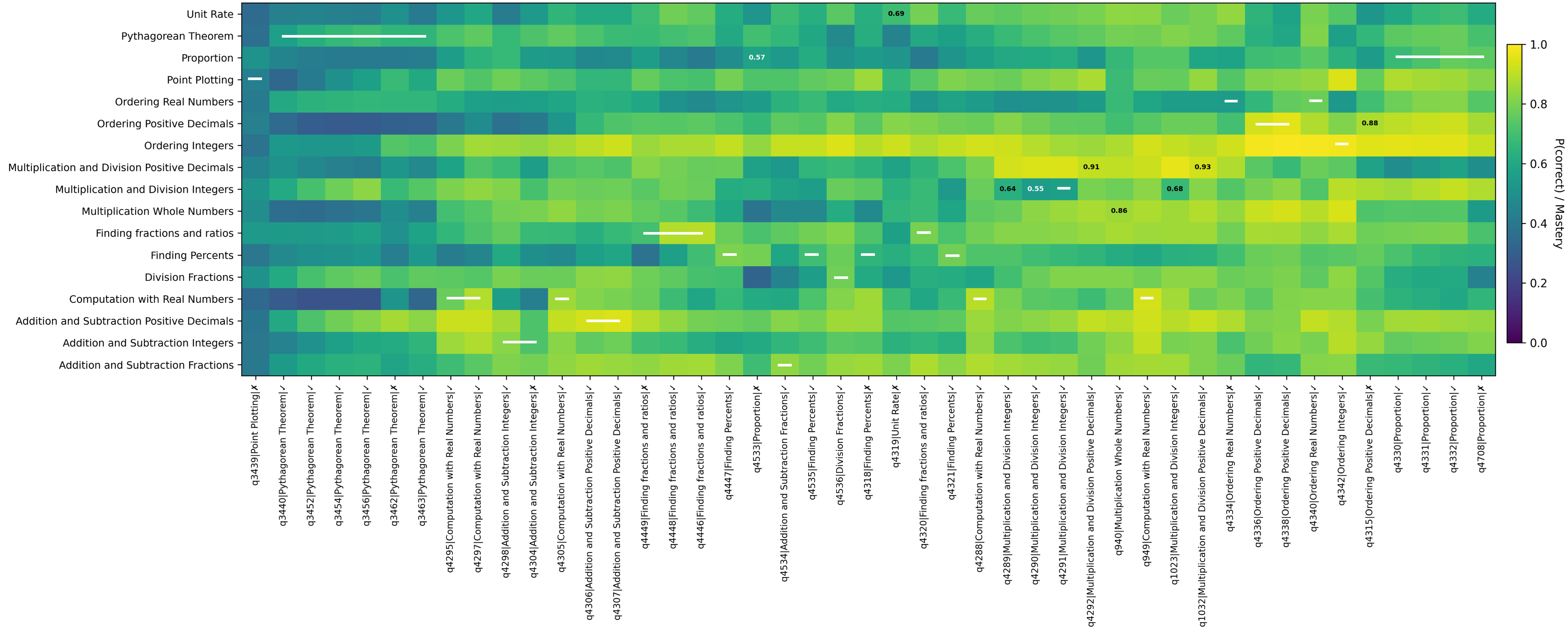}\\[-0.4em]
  {\normalsize (a)}\\[0.6em]

  \includegraphics[width=0.85\linewidth]{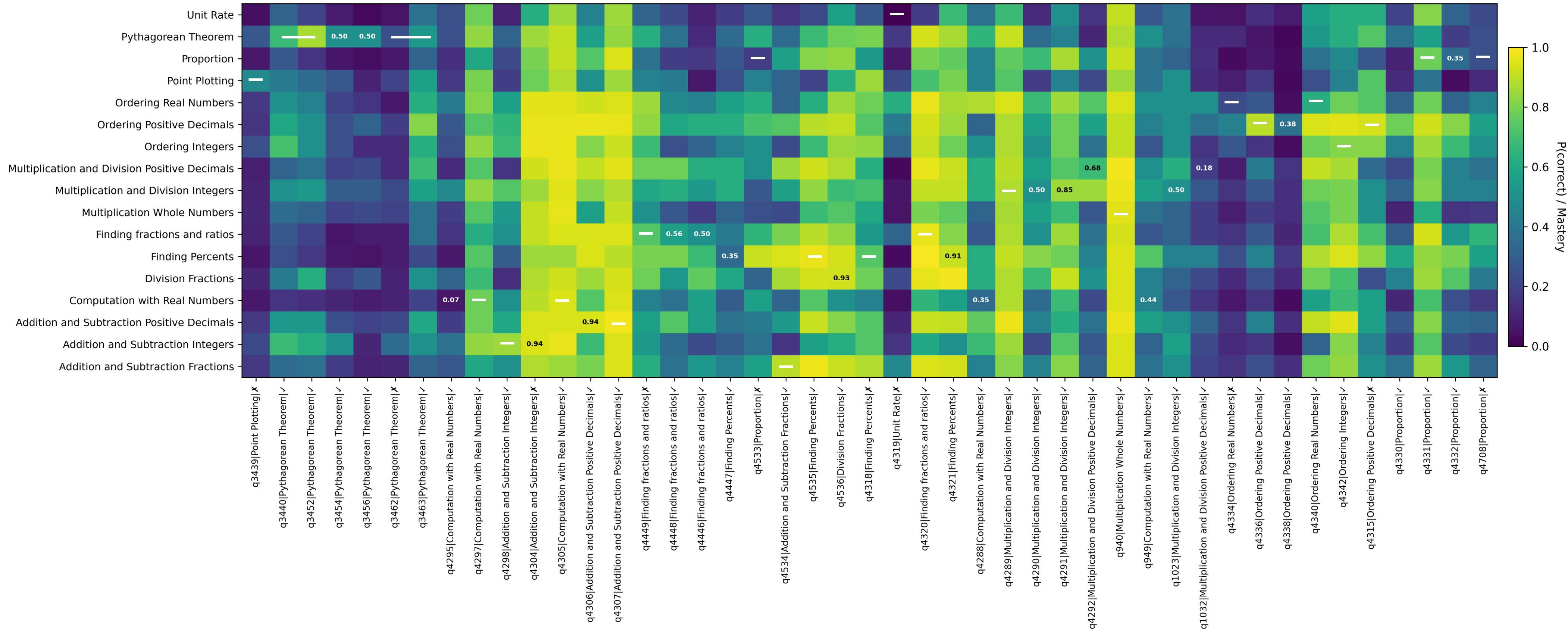}\\[-0.4em]
  {\normalsize (b)}

  \caption{Heatmaps of predicted skill-mastery probabilities for the test-set student with ID 81439: (a) DKT model, and (b) fine-tuned LLM model. Annotated cell values highlight points where the model’s mastery update was incorrect, while the white reference line represents the correct, temporally stable update trajectory for each skill—indicating how mastery should rise or fall based on the learner’s previous state and most recent response.}
  \label{fig:myfig5}
\end{figure}

\end{landscape}
\clearpage

As Fig~\ref{fig:myfig5}a shows, only a small number of mastery probabilities deviate from what would be expected based on the model’s sequential stability. A closer look shows two key points. First, there is only one instance in the early half of the sequence where a probability increases slightly when it should have decreased, given the student’s recent performance. Early-phase inaccuracies are important because they can lead to misleading instructional responses, especially if the model is used to provide real-time guidance. Second, across all 45-time steps, only nine inconsistent probabilities were observed, and most of these deviations were small and educationally negligible. All other cells without annotations represent predictions consistent with stable knowledge trajectories. For example, in sequences 2 to 7, the student repeatedly answered Pythagorean theorem items correctly. The DKT model gradually increased the predicted mastery after each correct response, then produced a slight, plausible decrease after an incorrect answer at sequence 6, before rising again after the next correct response. This pattern reflects the model’s sequential stability, showing smooth, plausible changes in knowledge that align with the student’s actual performance. 

Fig~\ref{fig:myfig5}b shows a different pattern for the fine-tuned LLM. Unlike the DKT model, the fine-tuned LLM exhibits less sequential stability. Nine inconsistent probabilities appear in the first half of the sequence alone. Such early fluctuations can distort initial mastery estimates and lead to poorly timed interventions. Across the full sequence, the fine-tuned model shows 19 inconsistencies overall, often producing abrupt probability shifts rather than smooth learning trajectories. For example, in sequences 28 to 30 for the “Multiplication and Division of Integers” skill, the student answered three items correctly, yet the fine-tuned model behaves inconsistently. It first increases the mastery estimate, then drops it to 50\% after another correct answer, and finally raises it again to 85\% at sequence 30. Although this final increase moves in the correct direction, the resulting probability is still lower than at sequence 28 and lower than what would be expected after three consecutive correct responses. This highlights that, even with fine-tuning, the model struggles to maintain coherent knowledge trajectories over time.

\section{Discussion and conclusions}

As generative AI becomes rapidly embedded in K–12 education, a central open question is whether general-purpose LLMs can responsibly fulfil the foundational requirement of adaptive tutoring—i.e., accurate and stable real-time assessment of learners’ learning trajectories. Motivated by this challenge, this study examines the extent to which LLMs can perform such assessment and compares their performance with a sequential learner modelling approach. By comparing DKT with both base and fine-tuned Llama 3-8B models on the ASSISTments 2009–2010 dataset, we assessed global prediction accuracy on the next-step correctness prediction task as well as the temporal coherence and sequential stability of their mastery updates. 

Across all quantitative and qualitative analyses, the DKT model consistently demonstrated superior accuracy, temporal stability, and pedagogical coherence compared with both zero-shot and fine-tuned LLMs. While fine-tuning Llama 3 8B improved next-step prediction performance relative to its zero-shot variant, the model still fell short of DKT, achieving lower AUC, less reliable behaviour across early–mid–late sequence segments, and substantially weaker temporal consistency. Volatility and inconsistency metrics further revealed that, despite learning to adjust mastery in the correct direction after fine-tuning, the LLM’s updates remained unstable and frequently contradicted observed evidence, whereas DKT maintained smooth, gradual, and pedagogically aligned mastery trajectories. Our multi-skill mastery estimation analysis reinforced this contrast: DKT produced coherent longitudinal representations of learning, while LLM outputs displayed rather abrupt fluctuations and a narrow, locally reactive pattern that failed to preserve temporal memory. These shortcomings were compounded by practical constraints: LLM inference and fine-tuning required substantially greater computational resources than DKT, which trained in minutes on modest hardware, thereby limiting their applicability in real-world educational settings.

Our findings align with an expanding body of evidence that raises concerns about the suitability of current LLMs for adaptive tutoring and learner modelling. Consistent with Borchers and Shou \cite{borchers2025can}, who show that even advanced models such as Llama 3-70B and GPT-4o struggle to reproduce the adaptivity of established ITSs, we likewise observe that LLMs fail to generate stable, instructionally meaningful mastery updates—specifically, probability shifts that correctly reflect a learner’s demonstrated performance over time. Their study further highlights that Llama 3-8B achieves the highest pedagogical quality among tested models while still violating basic instruction-following requirements. Our work extends the evidence base by focusing specifically on knowledge tracing: rather than evaluating instructional moves, we benchmark LLMs directly against DKT on next-step prediction, temporal dynamics, and multi-skill mastery trajectories, showing that even after computationally intensive fine-tuning on training dataset, the LLM remains temporally unstable and pedagogically inconsistent. This conclusion aligns with Neshaei et al. \cite{neshaei2024towards}, who similarly report that LLM-based approaches capture general interaction patterns yet fall short of specialised KT models. However, our study addresses limitations in their work by fine-tuning directly on numerical interaction sequences (rather than natural-language feature prompts) and by evaluating temporal coherence, volatility, inconsistency, sequence-stage errors, and cross-skill mastery predictions.  

A related contrast emerges when considering Li et al. \cite{li2025explainable}, who evaluate LLMs within an explainable few-shot KT framework. Although their study shows that LLMs such as GLM-4 and GPT-4 can match deep KT baselines on single-step prediction when supplied with structured textual inputs and rich semantic context, their formulation does not require the model to maintain temporally coherent mastery updates across full interaction sequences and relies heavily on semantically rich textual data. In contrast, our evaluation targets the core learner modelling requirement using only numerical interaction sequences: accurate next-step prediction combined with stable, sequential knowledge updating, assessed through temporal error patterns, volatility and inconsistency metrics, and multi-step mastery trajectories. Under these sequence-level and non-textual conditions, LLMs remain inconsistent and volatility-prone even after fine-tuning, whereas DKT preserves smooth, cumulative, and instructionally grounded learning trajectories. Moreover, the “explainability” promised by few-shot prompting does not address the fundamental opacity of LLMs: like other sub-symbolic deep models, they remain black boxes with billions of uninterpretable parameters, and their natural-language explanations reflect plausible surface-level reasoning rather than faithful accounts of internal decision processes \cite{hooshyar2025responsibleAI,lee2024life}. Compared with Lee et al. \cite{lee2024language} and Fu et al. \cite{fu2024sinkt}, who rely on textual semantics, language models trained to process rich contextual information, or heterogeneous concept–question graphs, our study isolates the ability of general LLMs to model learning purely from numerical interaction sequences. The results show that these models do not meet the temporal demands of learner modelling methods such as DKT. Viewed through the lens of Cho et al.’s \cite{cho2024systematic} research, our findings further reinforce concerns regarding the heavy computational burden of LLM-based KT. Our LLM required far greater computational resources than DKT and still produced erratic, pedagogically incorrect mastery trajectories. Finally, in relation to Park and Kim’s \cite{park2025comprehensive} study, our results empirically challenge the feasibility of LLM-standalone KT systems and highlight the practical limitations of current generation decoder-only LLMs: despite large-scale fine-tuning, the LLM failed to generate coherent, temporally consistent mastery estimates, whereas DKT remained robust, efficient, and pedagogically aligned. Taken together, this comparison demonstrates that the LLM examined in this study—whether zero-shot or fine-tuned—do not satisfy the foundational requirement of adaptive tutoring: accurate, stable, real-time assessment of learners’ evolving knowledge states. 

Beyond the comparative performance results, these findings also highlight a broader implication for the design of AI-driven educational systems: LLMs and learner modelling methods play fundamentally different roles and should not be viewed as interchangeable. Learner models remain crucial for tracing, predicting, and interpreting learners’ evolving cognitive and non-cognitive states, grounded in established learning theories and validated constructs. LLMs, by contrast, excel at generative tasks such as producing explanations, examples, or personalised feedback but do not provide the stable temporally grounded, evidence-based assessment required for adaptive decision-making. This distinction echoes the broader view in recent research that real-time knowledge tracing still requires purpose-built sequential models rather than general-purpose language models \cite{cho2024systematic}. 

In this light, \textbf{the future of AI in education is not a replacement of learner modelling with LLMs but a hybrid integration in which each component compensates for the other's limitations}. Responsible learner models can supply reliable, real-time assessments of mastery, misconceptions, practice effects, and learning trajectories, while LLMs can leverage that information to generate targeted explanations, adaptive practice items, motivational messages, or scaffolding aligned with the learner’s needs. Such systems would be proactive and context-aware, grounded in educational evidence yet enriched by the generative flexibility of modern LLMs. Hybrid architectures represent a promising direction for such integration. One pathway involves embedding representations from traditional KT models (e.g., DKT or BKT) directly into the representational space of LLMs, allowing sequence-aware learning dynamics to be fused with semantic-rich contextual reasoning \cite{wang2025llm}. Another complementary direction is the use of retrieval-augmented generation (RAG)-style designs, in which the temporally precise knowledge state tracked by the KT model becomes the retrieval layer that conditions LLM reasoning. In this configuration, the KT model functions as an authoritative source of mastery evidence, and the LLM generates pedagogically coherent feedback grounded in that evidence—combining accurate knowledge tracing with well-formed instructional guidance. Such retrieval-augmented hybrids have the practical advantage of reducing hallucinations by anchoring LLM outputs in verified educational signals rather than unconstrained generation \cite{arslan2024survey,fan2024survey}. A recent work conducted by Venugopalan et al. \cite{venugopalan2025combining} on hybrid tutoring shows how conversational recommendations generated by an LLM, when combined with learner modelling-based (tutoring) system intelligence, can support caregivers in providing content-level and metacognitive guidance to students, illustrating the value of human–AI integration in real learning contexts. Scarlatos et al. \cite{scarlatos2025exploring} demonstrate another hybrid pathway in which LLMs contribute to the tutoring pipeline not by performing learner modelling themselves, but by annotating dialogue data so that specialised KT models can carry out the actual knowledge tracing task—further reinforcing the complementary roles that generative models and learner modelling methods can play in a unified system. Similarly, Wang et al. \cite{wang2025llm} propose a hybrid LLM-KT framework that aligns LLMs with traditional sequence models using plug-in instructions, context encoders, and sequence adapters. This approach explicitly combines the semantic strengths of LLMs with the temporal precision of established KT architectures. However, realising this vision requires hybrid human–AI based learner modelling (codesigned with stakeholders to meet requirements for human-centeredness, ethical and reliable decision-making, and methodological explainability), where generative AI and human expertise are integrated in a coherent framework \cite{hooshyar2025responsibleAI}. Human-in-the-loop design ensures that pedagogical principles guide model development and that outputs remain interpretable and accountable. Methods such as neural-symbolic computing illustrate this pathway. In practice, symbolic knowledge from educational contexts can be embedded into AI systems and refined through continuous use, strengthening both technical validity and pedagogical relevance \cite{tato2022infusing}. Another example of such a method is the use of neural-symbolic architectures that combine rule-based reasoning with LLMs, enabling systems to generate recommendations that remain responsible and compliant with pedagogical constraints. Taken together, these perspectives reinforce the implications of our empirical findings: \textit{while general-purpose LLMs alone are not suitable as core learner modelling models, their integration with specialised learner models, supported by human pedagogical oversight, offers a viable and responsible path forward for building adaptive, transparent, and instructionally meaningful educational AI systems}.

\subsection{Limitations and future works}

This study focused on one foundational dimension of responsible AI in education, accurate and reliable decision making, because it underpins responsible tutoring in high-risk contexts such as K–12 education, as recognised by the EU AI Act. We evaluated an open source LLM on next-step correctness prediction and multi-skill mastery estimation, two core tasks in knowledge tracing. Other essential dimensions of responsible AI, including human-centred design, ethical decision making, user trust, explainability, and privacy-preserving implementation, were beyond the scope of this work. Future research should extend our findings by exploring learner modelling methods that support these broader dimensions, such as hybrid neural-symbolic approaches for DKT. Such methods could offer greater transparency, reduced bias, and improved human-centredness compared with purely data-driven models, which can be prone to learning biases, limited interpretability, and weaker alignment with responsible AI principles. A second limitation concerns the fine-tuning strategy adopted for the LLM. We employed a parameter-efficient LoRA configuration rather than full-model fine-tuning, in line with common practice for resource-constrained and reproducible experimentation. While this approach allows controlled adaptation without overwriting base representations, it may not fully exploit the model’s capacity. Future work should systematically examine how fine-tuning depth and adaptation strategies influence LLM performance in learner modelling. A further limitation concerns model coverage: our analysis evaluated only the Llama3-8B from Llama family of models. Although selected for their strong performance in tutoring-related tasks, future work should compare a wider range of LLM architectures and scales to assess the generalisability of our conclusions. Finally, we relied on a single large-scale dataset. Replicating the study across diverse datasets would further validate the findings and clarify when and how LLMs can be responsibly integrated within K--12 education.

\section*{Acknowledgements}

This work was supported by the Estonian Research Council grant (PRG2215).

\bibliographystyle{unsrt}
\bibliography{references}

\end{document}